%% file: paper.tex
\documentclass[10pt,twocolumn,letterpaper]{article}

\usepackage{cvpr}              

\usepackage{graphicx}
\usepackage{amsmath}
\usepackage{amssymb}
\usepackage{booktabs}
\usepackage{multirow}
\usepackage[font=small,skip=1pt]{caption}
\usepackage{pifont}
\newcommand{\cmark}{\ding{51}}%
\usepackage{bm}
\usepackage[accsupp]{axessibility} 

\usepackage[pagebackref,breaklinks,colorlinks]{hyperref}

\usepackage[capitalize]{cleveref}
\crefname{section}{Sec.}{Secs.}
\Crefname{section}{Section}{Sections}
\Crefname{table}{Table}{Tables}
\crefname{table}{Tab.}{Tabs.}

\def\cvprPaperID{4821} 
\def\confName{CVPR}
\def\confYear{2022}

\begin{document}

\title{PatchNet: A Simple Face Anti-Spoofing Framework \\ 
via Fine-Grained Patch Recognition}

\author{ 
    Chien-Yi Wang\textsuperscript{1} \qquad 
    Yu-Ding Lu\textsuperscript{2} \qquad 
    Shang-Ta Yang\textsuperscript{1} \qquad 
    Shang-Hong Lai\textsuperscript{1} \qquad 
    \\
    \textsuperscript{1}Microsoft AI R\&D Center, Taiwan
    \quad
    \textsuperscript{2}HTC
    \\
    {\tt\small \{chiwa, shanya, shlai\}@microsoft.com}
    \quad
    {\tt\small jonlu.citi@gmail.com}
} 
\maketitle

\begin{abstract}
   Face anti-spoofing (FAS) plays a critical role in securing face recognition systems from different presentation attacks. Previous works leverage auxiliary pixel-level supervision and domain generalization approaches to address unseen spoof types. However, the local characteristics of image captures, i.e., capturing devices and presenting materials, are ignored in existing works and we argue that such information is required for networks to discriminate between live and spoof images.
   In this work, we propose PatchNet which reformulates face anti-spoofing as a fine-grained patch-type recognition problem. To be specific, our framework recognizes the combination of capturing devices and presenting materials based on the patches cropped from non-distorted face images. This reformulation can largely improve the data variation and enforce the network to learn discriminative feature from local capture patterns. In addition, to further improve the generalization ability of the spoof feature, we propose the novel Asymmetric Margin-based Classification Loss and Self-supervised Similarity Loss to regularize the patch embedding space.
   Our experimental results verify our assumption and show that the model is capable of recognizing unseen spoof types robustly by only looking at local regions. Moreover, the fine-grained and patch-level reformulation of FAS outperforms the existing approaches on intra-dataset, cross-dataset, and domain generalization benchmarks. Furthermore, our PatchNet framework can enable practical applications like Few-Shot Reference-based FAS and facilitate future exploration of spoof-related intrinsic cues.
\end{abstract}

\input{1_introduction.tex}
\input{2_related_works.tex}
\input{3_proposed_methods.tex}
\input{4_experiments.tex}

\input{5_conclusions.tex}

{\small
\bibliographystyle{ieee_fullname}
\bibliography{egbib}
}

\end{document}

%% file: 1_introduction.tex
\section{Introduction}
Face anti-spoofing (FAS) is a crucial technique to prevent face recognition systems from security attacks. With the advance of deep neural network, several learning-based approaches were proposed to discriminate live faces from physical presentation attacks.

\begin{figure}[t]
    \centering
    \includegraphics[width=.95\linewidth]{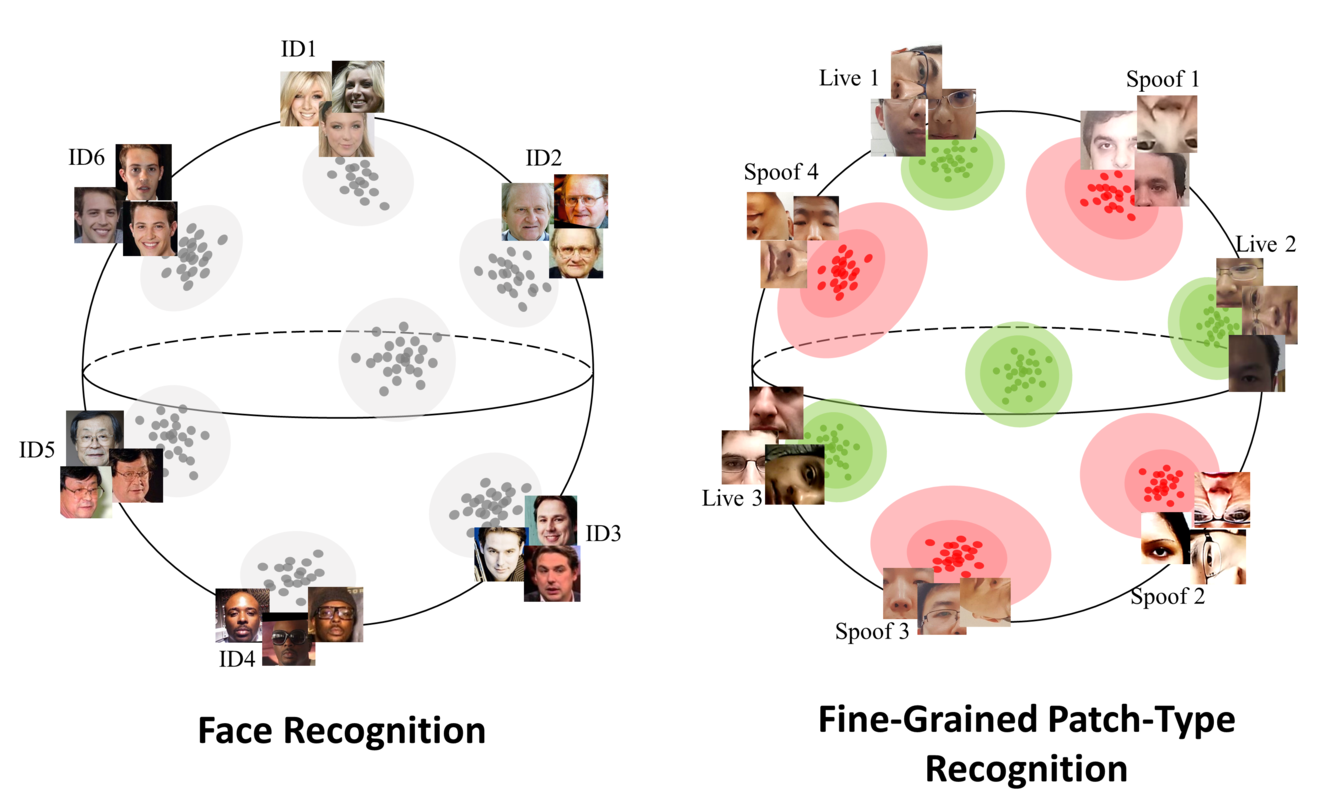}
    \caption{The face recognition model learns a face embedding space to discriminate between identities. Our \textbf{fine-grained patch-type recognition model} learns a patch embedding space to discriminate between patches with different capture characteristics.}
    \label{fig:teaser}
    \vspace{-6.0mm}
\end{figure}

Previous face anti-spoofing methods are highly limited by the scale and variation of the datasets. Commonly used datasets~\cite{boulkenafet2017oulu, liu2018learning, wen2015msumfsd, chingovska2012replayattack, zhang2012casia} contain less than 100 identities during training, and the spoof images are captured under limited variation. Based on our observation, training on such datasets with a binary classification model is prone to overfit to the biases introduced by the data collection, and the learned features are vulnerable in the unseen testing scenarios. Therefore, previous face anti-spoofing works~\cite{liu2018learning, shao2020regularized, yu2020face, yu2021dual, liu2021dual, liu2021adaptive} leverage auxiliary pixel-wise supervision (e.g., the facial depth map and reflection map) as a strong prior knowledge to achieve better generalization ability under testing scenarios with unseen illumination or spoof types. The other FAS works~\cite{liu2020disentangling, zhang2020face} propose to adopt Generative Adversarial Network (GAN) to disentangle the feature maps of live faces and spoof images by reconstructing new live and spoof facial images. Despite the effectiveness of these spoof-detecting techniques, it is still remained as an open question to describe the intrinsic cues learned from networks. Yu~\etal ~\cite{yu2020face} rephrase FAS as a structural material recognition problem, which assumes that the discrimination of the structural materials between human facial skin and physical spoofing carriers is the essence for FAS tasks. Following the similar motivation, we believe that the capability of recognizing and comparing different fine-grained material types is the key to learn robust intrinsic cues for FAS.

\begin{figure*}[t]
    \centering
    \includegraphics[width=.9\linewidth]{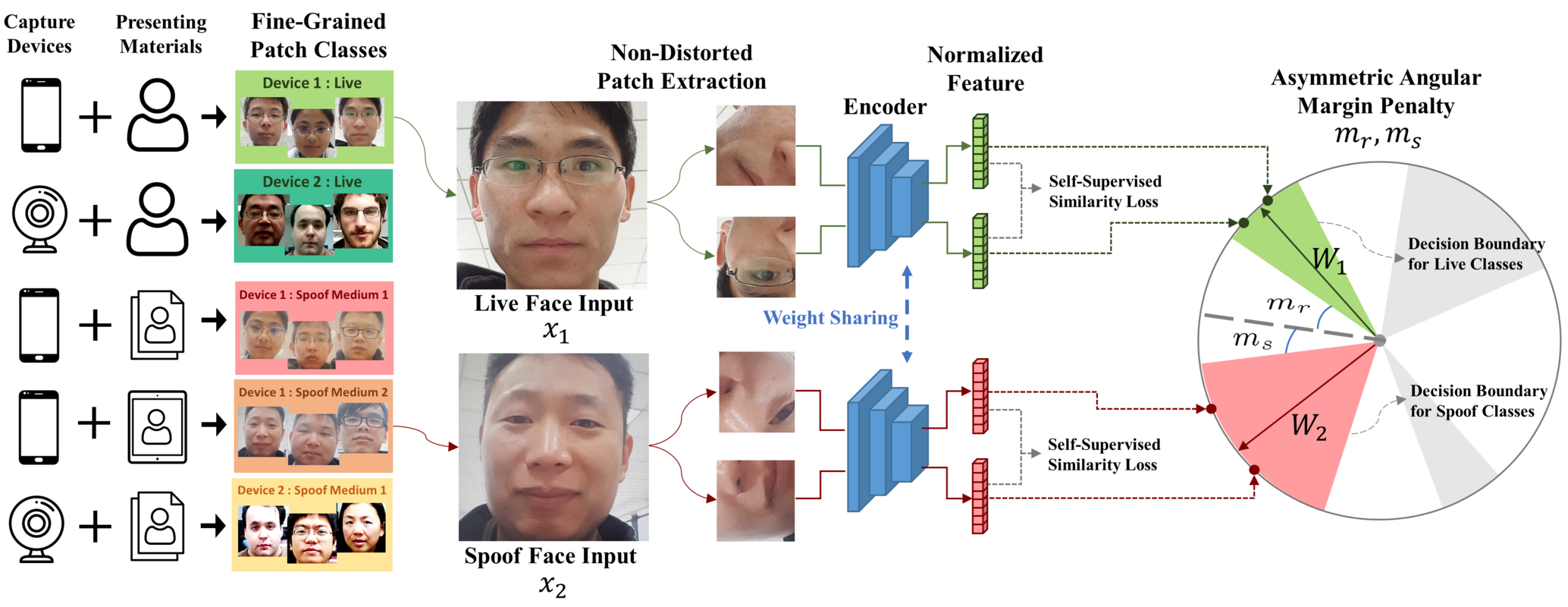}
    \caption{\textbf{Overview of our proposed PatchNet framework}. We address the face anti-spoofing with a fine-grained patch-type recognition model. The patch-type classes are pre-defined by the combination of the capture device and the presenting material, and the patch inputs are extracted from the face captures by non-distorted augmentation operations. \textbf{Asymmetric Angular Margin Softmax Loss} is employed in the last classification layer to impose larger angular margin on live classes. \textbf{Self-Supervised Similarity Loss} is applied to enforce the patch feature invariance within a single capture.}
    \label{fig:overview}
    \vspace{-4.0mm}
\end{figure*}

In this paper, we propose PatchNet which learns discriminative features based on patches cropped from the entire face regions. Inspired by previous works~\cite{yang2019face, cai2021drlfas, deb2020look}, the patch-level inputs can enhance the data variation and enforce the network to learn spoof-specific features in the local region, and thus prevent the network from overfitting to the biases introduced by datasets. Instead of resizing the input face images into the same size as adopted by recent FAS works, we directly crop the fixed-size patches from raw facial captures to avoid the distortion of discriminative FAS cues. With the patch-level inputs, our PatchNet aims at classifying the corresponding fine-grained categories, i.e., the capturing devices and presenting materials, and we denote each category as a specific ``patch-type". To enforce the network to learn robust spoof-related feature to recognize unseen patch types during testing, we adopt the angular margin-based softmax loss that is commonly used in face recognition tasks~\cite{wang2018additive, deng2019arcface, wang2018cosface}, which aims to optimize the face embedding on the normalized hypersphere (Figure~\ref{fig:teaser}). Moreover, since the patch type classes are not symmetric between live and spoof faces, we propose ``asymmetric angular margin loss'' and impose a larger margin on live type classes. Inspired by the recent works on self-supervised learning~\cite{chen2020simple, chen2020exploring, grill2020bootstrap}, and the fact that material patterns are presented spatially in the entire face region, we also propose ``self-supervised similarity loss'' to regularize the features with location and rotation invariance.

To demonstrate the effectiveness of PatchNet, we conduct extensive experiments on intra-dataset, cross-dataset, and domain generalization benchmark datasets, and PatchNet achieves the state-of-the-art performance under most testing scenarios. Moreover, we also conduct the ablation study to further investigate the proposed components. 

Our contributions are summarized as follows:

\begin{itemize}
    \item We reformulate face anti-spoofing as a fine-grained patch recognition problem, and design a simple framework called PatchNet to learn an embedding space to encode intrinsic cues from local patches to represent captures' characteristics.
    \item We propose novel Asymmetric Margin-based Softmax Loss and Self-supervised Similarity Loss to supervise the PatchNet training. While the former helps to learn a more generalized patch type embedding space to address the asymmetry between live and spoof, the latter can enforce the patch feature to be invariant within a single capture.
    \item The proposed framework could achieve state-of-the-art performance on intra-dataset, cross-dataset, and domain generalization benchmarks simultaneously without auxiliary pixel-wise supervision and domain generalization techniques. Moreover, the learned patch embedding space can enable applications like Few-Shot reference FAS and patch type retrieval, which can boost the FAS performance in certain deployment scenarios.
\end{itemize}

%% file: 2_related_works.tex
\section{Related Works}
\textbf{Auxiliary-based Methods.} Most of the recent works leverage auxiliary tasks as the prior knowledge to guide the feature learning toward more generalizable cues. Liu \etal~\cite{liu2018learning} proposed to employ the depth map and rPPG as strong supervision signals for live samples to regularize the features. Kim \etal~\cite{kim2019basn} further leveraged the reflection maps as the supervision signal for spoof samples. Many other FAS methods~\cite{shao2020regularized, yu2020face, feng2020learning, shao2019multi, yu2021dual, liu2021dual, liu2021adaptive} also heavily rely on similar auxiliary pixel-wise supervision to improve their FAS model performance. Even though the feature learning can benefit from such supervision, the pseudo ground truths for those tasks are not accurate, and the generation of those supervision signals takes high computation resources. 


\textbf{Domain Generalization FAS Methods.} In the face anti-spoofing community, domain generalization techniques are developed to address the domain shift between different anti-spoofing datasets. Shao \etal~\cite{shao2020regularized} employed meta-learning techniques to simulate the target domain shift during the training process to regularize the feature learning directions. Wang \etal~\cite{wang2020cross} proposed to learn domain-independent features via a disentangled representation learning framework. The most related work to ours is ~\cite{jia2020single}, which treats live and spoof samples asymmetrically and applies adversarial loss and triplet loss to regularize the features in the normalized space. Actually, domains are hard to define in FAS tasks, as even within the same dataset, there are captures with very different capture devices. While people are using generalization methods to find common features across collections and spoof types, we aim to break the concept of domain and propose to learn a generic embedding space that encodes capture characteristics explicitly.

%% file: 3_proposed_methods.tex
\section{Proposed Method}
\subsection{Overview}
As illustrated in Fig.~\ref{fig:overview}, we reformulate face anti-spoofing as a fine-grained patch-type recognition problem and propose a simple training framework to learn the patch features efficiently. First, we apply certain transform on the original image to obtain the patch inputs, and the patch features are extracted by an encoder and then normalized in the feature space. Based on the meta-info from the training dataset, we split the categories finely based on the \textbf{presenting materials} and \textbf{capture devices}. For example, in CASIA-FASD, there are two different spoof mediums and three different capture resolutions, so there are nine fine-grained patch types (three live and six spoof types).  

Inspired by the latest face recognition approaches, during training we employ angular margin-based softmax loss, which can force feature cluster for each category to be compactly distributed and enable better generalization ability. Furthermore, as the distribution discrepancies between spoof samples are larger than live samples, we treat live and spoof samples asymmetrically: force the model to learn a more compact cluster within live samples while leaving spoof samples more dispersed in the feature space. We modify the angular margin-based softmax loss and apply asymmetric margin onto live and spoof patch types: imposing larger angular margin on live types to push more compact boundaries. Finally, the self-supervised similarity loss further regularizes the patch features by applying the positive part of the contrastive loss on two transformed patch views from a single whole face image. Given that spoof-specific discriminative information is present spatially in the entire face region, the features between two different patch views from the same face capture should be similar. 

\subsection{Patch Features Extraction}
We want to avoid any transform which can lead to image distortion or the reduction of the important spoof-related information. Given the cropped face region $x_i$ from the raw capture, the two augmented patch views from $x_i$ are $x_i^{t_1} = t_1(x_i)$ and $x_i^{t_2} = t_2(x_i)$, where $t_1, t_2 \sim \mathcal{T}$. $\mathcal{T}$ is the sequence of non-distorted augmentation operations, which only have \textbf{random horizontal flip}, \textbf{random rotation}, and \textbf{fixed size cropping}. The two input patches are then passed into the encoder $E_\theta$ and normalization layer to get the final features: $f_i^{t_1} = Normalize(E_\theta(x_i^{t_1})), f_i^{t_2} = Normalize(E_\theta(x_i^{t_2}))$.   
\subsection{Fine-Grained Patch Recognition}
Assuming we have $N$ patch type classes in the training dataset, which consists of $k$ live and $N-k$ spoof classes. Each input patch $t(x_i)$ belongs to one fine-grained ground truth class $y_i \in \{L_1, L_2, ... L_k, S_1, S_2, ... S_{N-k}\}$, and the Angular-Margin Softmax Loss is applied to regularize the patch features. The Angular-Margin Softmax Loss has many variants~\cite{wang2018additive, deng2019arcface, liu2016large, wang2018cosface, liu2017sphereface} and is commonly used in face recognition to improve the generalization ability to open-set identities. In this work, we employ AM-Softmax~\cite{wang2018additive} loss to optimize the fine-grained patch recognition model and modify it to address the asymmetric nature in face anti-spoofing.
\subsubsection{Preliminaries}
The formulation of the original Softmax loss is given by
\begin{equation}
\begin{aligned}
\mathcal{L}_S & = -\frac{1}{n}\sum_{i=1}^n{log\frac{e^{W_{y_i}^T \bm{f}_i}}{\sum_{j=1}^{c}{e^{W_j^T \bm{f}_i}}}}\\
& = -\frac{1}{n}\sum_{i=1}^n{log\frac{e^{\|W_{y_i}\| \|\bm{f}_i\|cos(\theta _{y_i})}}{\sum_{j=1}^{c}{e^{\|W_j\| \|\bm{f}_i\| cos(\theta_{j})}}}},
\label{eq:softmax}
\end{aligned}
\end{equation}

where $\bm{f}$ is the input of the fully connected layer for classification ($\bm{f}_i$ denotes the $i$-th sample), $W_j$ is the $j$-th column of the fully connected layer, and $y_i$ is the ground truth label of the $i$-th sample. The term $W_{y_i}^T \bm{f}_i$ is also called the target logit of the $i$-th sample.

The large-margin property is introduced by Sphereface~\cite{liu2017sphereface}, which defines a general function $\psi(\theta)$ to impose the angular margin between feature and weight vectors. After applying feature and weight normalization ($\|W_{y_i}\| = \|\bm{f}_i\| = 1$), the loss function becomes
\begin{equation}
\mathcal{L}_{S}  = -\frac{1}{n}\sum_{i=1}^n{log\frac{e^{\psi(\theta _{y_i})}}{e^{\psi(\theta _{y_i})} + \sum_{j=1,j\neq y_i}^{c}{e^{cos(\theta_{j})}}}},
\label{eq:a-softmax}
\end{equation}
where in AM-Softmax~\cite{wang2018additive} the function $\psi(\theta)$ is defined as
\begin{equation}
\psi(\theta) = cos\theta - m
\label{eq:psi_hardmargin}
\end{equation}

During implementation, the input after normalizing both the feature and the weight is actually $x = cos\theta_{y_i} = \frac{W_{y_i}^T f_i}{\|W_{y_i}\| \|f_i\|}$, so in the forward propagation it only needs to compute
\begin{equation}
\begin{aligned}
\Psi(x) &= x - m
\end{aligned}
\label{eq:psi_on_cos}
\end{equation}

Then it scales the cosine values using a hyper-parameter $s$ and the final AM-Softmax loss function becomes
\begin{equation}
\begin{aligned}
\mathcal{L}_{AMS} & = -\frac{1}{n}\sum_{i=1}^n{log\frac{e^{s \cdot  \left(cos\theta_{y_i} - m \right)}}{e^{s \cdot \left(cos\theta_{y_i} - m \right)} + \sum_{j=1,j\neq y_i}^{c}{e^{s \cdot  cos\theta_{j}}}}}\\
& = -\frac{1}{n}\sum_{i=1}^n{log\frac{e^{s \cdot  \left(W_{y_i}^T \bm{f}_i - m \right)}}{e^{s \cdot  \left(W_{y_i}^T \bm{f}_i - m \right)} + \sum_{j=1,j\neq y_i}^c{e^{s W_j^T \bm{f}_i}}}}.
\end{aligned}
\label{eq:am-softmax}
\end{equation}

\subsubsection{Asymmetric AM-Softmax Loss}
We impose different angular margin $m_l$ and $m_s$ on live and spoof categories, respectively. Denote live category set as $L = \{L_1, L_2, ... L_k\}$ and spoof category set as $S = \{S_1, S_2, ... S_{N-k}\}$. The modified AM-Softmax Loss of one feature sample $f_i$ becomes
\begin{equation}
\begin{aligned}
\mathcal{L}_{AAMS}(\bm{f}_i) & = \begin{cases}
-log\frac{e^{s \cdot  \left(W_{y_i}^T \bm{f}_i - m_l \right)}}{e^{s \cdot \left(W_{y_i}^T \bm{f}_i - m_l \right)} + \sum\limits_{j=1,j\neq y_i}^{N}{e^{s \cdot  W_j^T \bm{f}_i}}}  & \text{${y_i}\in{L}$}\\
-log\frac{e^{s \cdot  \left(W_{y_i}^T \bm{f}_i - m_s \right)}}{e^{s \cdot \left(W_{y_i}^T \bm{f}_i - m_s \right)} + \sum\limits_{j=1,j\neq y_i}^{N}{e^{s \cdot  W_j^T \bm{f}_i}}} & \text{${y_i}\in{S}$}
\end{cases} 
\end{aligned}
\label{eq:asym-am-softmax}
\end{equation}

The final Asymmetric Recognition Loss on two augmented patch views from the image is formulated as
\begin{equation}
\begin{aligned}
\mathcal{L}_{Asym} = -\frac{1}{n}\sum_{i=1}^n (\mathcal{L}_{AAMS}(\bm{f}_i^{t_1}) + \mathcal{L}_{AAMS}(\bm{f}_i^{t_2}))
\end{aligned}
\label{eq:final-asym-loss}
\end{equation}

\subsection{Self-Supervised Similarity Loss}
Given two different patch views from the same face image, the self-supervised similarity constraint is applied to enforce the features to be similar. Therefore, the spoof-related feature can be learned with patch location and rotation invariance.
\begin{equation}
\begin{aligned}
\mathcal{L}_{Sim}(\bm{f}_i^{t_1}, \bm{f}_i^{t_2}) & = \frac{1}{n}\sum_{i=1}^n \left\|\bm{f}_i^{t_1} - \bm{f}_i^{t_2}\right\|_2
\end{aligned}
\label{eq:sim-loss}
\end{equation}

\subsection{Training and Testing}
\subsubsection{Total Loss}
The total loss $L$ of the proposed framework during training is
\begin{equation}
\begin{aligned}
\mathcal{L} = \alpha_{1} \mathcal{L}_{Asym} + \alpha_{2} \mathcal{L}_{Sim}
\end{aligned}
\label{eq:total-loss}
\end{equation}
where $\alpha_{1}$ and $\alpha_{2}$  are the weights to balance the influence of loss components. In all of the experiments, we set $\alpha_{1} = \alpha_{2} = 1.0$.
\subsubsection{Testing Strategy}
Given a test face image, we uniformly crop patches from the whole image for the network inference, with the patch size the same as the one in the training process. Assuming we have $P$ cropped patches features ($f^1, f^2, ... f^P$) from one face image, then the average live probability can be obtained by the sum of live class probabilities in the last fully connected layer: 
\begin{equation}
\begin{aligned}
LiveProb = \frac{1}{P} \sum_{i=1}^P \sum_{y \in L}Softmax(s \cdot W_{y}^T \bm{f^i}) 
\end{aligned}
\label{eq:spoof-score}
\end{equation}

%% file: 4_experiments.tex
\section{Experiments}
\subsection{Datasets and Protocols} \label{sec:dataset}
\textbf{Databases.} Five databases OULU-NPU~\cite{boulkenafet2017oulu} (denoted as O), SiW~\cite{liu2018learning} (denoted as S), CASIA-FASD~\cite{zhang2012casia} (denoted as C), Replay-Attack~\cite{chingovska2012replayattack} (denoted as I), MSU-MFSD~\cite{wen2015msumfsd} (denoted as M) are used in the testing protocols. OULU-NPU and SiW are large-scale high-resolution databases containing four and three protocols to validate the generalization (e.g., unseen environment and spoof mediums) of models, respectively, which are utilized for intra-dataset testing. CASIA-MFSD, Replay-Attack, and MSU-MFSD are databases that contain low-resolution videos with much fewer video clips and are used for cross-dataset testing to validate the generalization ability to testing data with large distribution shift. There are three capture devices with quality ranging from low to high in CASIA-FASD, two devices in SiW and MSU-MFSD, and only one device in the other datasets. The fine-grained class number and the other statistics of databases are shown in Tab.~\ref{tab:database}. Note that in Oulu-NPU, even the collections are captured by six different types of phones, the quality and fine details are pretty similar, so we only split the patch type into five classes in total. More details and sample images can be found in the supplementary material.
\begin{table}[t]
    \scriptsize
    \centering
    \resizebox{0.95\columnwidth}{!}{\begin{tabular}{c|c|c|c|c|c|c}
        \toprule
        \multirow{2}*{\textbf{Dataset}} & \multicolumn{2}{c|}{\textbf{\# Subjects}} & \multicolumn{2}{c|}{\textbf{\# Clips}} & 
        \multicolumn{2}{c}{\textbf{\# Classes}} \\
        \cline{2-7}
        & Train & Test & Train & Test & Live & Spoof \\
        \hline
        \hline
        OULU-NPU (O) & 20 & 20 & 1800 & 1800 & 1 & 4 \\
        \hline
        SiW (S) & 90 & 75 & 2442 & 2036 & 2 & 12 \\
        \hline
        CASIA-FASD (C) & 20 & 30 & 480 & 720 & 3 & 6 \\
        \hline
        ReplayAttack (I) & 30 & 20 & 360 & 240 & 1 & 3 \\
        \hline
        MSU-MFSD (M) & 15 & 20 & 120 & 160 & 2 & 6 \\
        \bottomrule
    \end{tabular}}
    \caption{Statistics of the face anti-spoofing datasets.}
    \label{tab:database}
    \vspace{-5.5mm}
\end{table}

\textbf{Performance Metrics.} In intra-dataset testing on OULU-NPU and SiW, we follow the original protocols and metrics, i.e., Attack Presentation Classification Error Rate (APCER), Bona Fide Presentation Classification Error Rate (BPCER), and Average Classification Error Rate (ACER) for a fair comparison. Half Total Error Rate (HTER) and Area Under Curve
(AUC) are adopted in the cross-dataset testing between OULU-NPU, CASIA-MFSD, Replay-Attack, and MSU-MFSD.

\subsection{Implementation Details} \label{sec:implementation}
All face anti-spoofing datasets above are stored in video format originally. We randomly select three frames from each video clip and use the state-of-the-art face detector RetinaFace~\cite{deng2020retina} to crop the face for training. We set the fixed patch crop size as 160, and set the hyperparameter $s = 30.0, m_l = 0.4, m_s = 0.1$ in all protocols. We use ResNet18~\cite{he2016deep} as the patch feature encoder, and we did not see much performance difference while using an encoder with larger capacity (as shown in the supplementary materials). Models are trained with SGD optimizer and the initial learning rate is 0.002. We train models with maximum 200 epochs while the learning rate halves every 90 epochs. During testing, we uniformly crop the fixed-size patches from the face input image: the minimum x and y coordinates are $size / 2.0$, and the maximum x and y coordinates are $width - (size / 2.0)$ and $height - (size / 2.0)$, respectively. In all of the experiments during testing, we uniformly sample 3 patch anchors on each side, which results in $P=9$ patches for score averaging.       

\subsection{Intra-Dataset Testing} \label{sec:intra-results}
We conduct experiments on Oulu-NPU~\cite{boulkenafet2017oulu} and SiW~\cite{liu2018learning} for intra-dataset testing results. We compare the results with the most recent face anti-spoofing methods in the following.
\subsubsection{Results on Oulu-NPU}
Oulu-NPU~\cite{boulkenafet2017oulu} has four challenging protocols, which evaluate the model robustness against the unseen environment, unseen spoof mediums, unseen capture devices, and all of the above, respectively. The number of classes during training are 5, 3, 5, and 3, respectively. As shown in Tab.~\ref{tab:OULU}, our simple patch-based recognition approach achieves the best performance in all protocols. It clearly verifies the better generalization ability of the features learned through the patch recognition proxy tasks.    
\begin{table}[t]
\centering
\resizebox{0.95\columnwidth}{!}{\begin{tabular}{c|c|c|c|c}
\toprule
Prot. & Method & APCER(\%) & BPCER(\%) & ACER(\%) \\
\hline
\hline
\multirow{7}{*}{1}
        &Disentangle ~\cite{zhang2020face}& 1.7& 0.8& 1.3\\
        &SpoofTrace ~\cite{liu2020disentangling} &0.8 &1.3 &1.1 \\
        &BCN ~\cite{yu2020face} & 0.0 & 1.6 & 0.8 \\
        &CDCN ~\cite{yu2020searching} &0.4 & 1.7 & 1.0 \\
        &NAS-FAS ~\cite{yu2020fas} &0.4 &0.0 & \underline{0.2} \\
        &\textbf{PatchNet (Ours)} & 0.0 & 0.0 & \textbf{0.0}\\
\hline
\multirow{7}{*}{2} 
       &Disentangle ~\cite{zhang2020face}& 1.1 & 3.6 & 2.4\\
       &SpoofTrace ~\cite{liu2020disentangling} & 2.3 & 1.6 & 1.9\\
       &BCN ~\cite{yu2020face} & 2.6 & 0.8 & 1.7 \\
       &CDCN ~\cite{yu2020searching} & 1.5 & 1.4 & \underline{1.5} \\
       &NAS-FAS ~\cite{yu2020fas} & 1.5 & 0.8 & \textbf{1.2} \\
       &\textbf{PatchNet (Ours)} & 1.1 & 1.2 & \textbf{1.2} \\
\hline
\multirow{7}{*}{3} 
       &Disentangle ~\cite{zhang2020face}& 2.8$\pm$2.2& 1.7$\pm$2.6 & 2.2$\pm$2.2\\
       &SpoofTrace ~\cite{liu2020disentangling} & 1.6$\pm$1.6& 4.0$\pm$5.4& 2.8$\pm$3.3\\
       &BCN ~\cite{yu2020face} & 2.8$\pm$2.4 & 2.3$\pm$2.8 & 2.5$\pm$1.1 \\
       &CDCN ~\cite{yu2020searching}& 2.4$\pm$1.3 &2.2$\pm$2.0 & 2.3$\pm$1.4 \\
       &NAS-FAS ~\cite{yu2020fas} &2.1$\pm$1.3 &1.4$\pm$1.1  & \underline{1.7$\pm$0.6} \\
       &\textbf{PatchNet (Ours)} & 1.8$\pm$1.47 & 0.56$\pm$1.24 & \textbf{1.18$\pm$1.26}\\
\hline
\multirow{7}{*}{4}
       &Disentangle ~\cite{zhang2020face}& 5.4$\pm$2.9 & 3.3$\pm$6.0& 4.4$\pm$3.0\\
       &SpoofTrace ~\cite{liu2020disentangling} &2.3$\pm$3.6& 5.2$\pm$5.4& 3.8$\pm$4.2\\
       &BCN ~\cite{yu2020face} & 2.9$\pm$4.0 & 7.5$\pm$6.9 & 5.2$\pm$3.7 \\
       &CDCN ~\cite{yu2020searching}&4.6$\pm$4.6 &9.2$\pm$8.0  & 6.9$\pm$2.9 \\
       &NAS-FAS ~\cite{yu2020fas} &4.2$\pm$5.3 &1.7$\pm$2.6 & \textbf{2.9$\pm$2.8} \\
       &\textbf{PatchNet (Ours)} &2.5$\pm$3.81 & 3.33$\pm$3.73& \underline{2.9$\pm$3.0}\\
\bottomrule
\end{tabular}}
\caption{The results of testing on OULU-NPU protocols.}
\label{tab:OULU}
\vspace{-4.5mm}
\end{table}

\begin{table}
\centering
\resizebox{0.95\columnwidth}{!}{\begin{tabular}{c|c|c|c|c}
\toprule
Prot. & Method & APCER(\%) & BPCER(\%) & ACER(\%) \\
\hline
\hline
\multirow{5}{*}{1} 
      &Disentangle ~\cite{zhang2020face}&0.07 &0.50 &0.28 \\
      &SpoofTrace ~\cite{liu2020disentangling} & 0.00 & 0.00& \textbf{0.00}\\
      &BCN ~\cite{yu2020face} & 0.55& 0.17& 0.36\\
      &CDCN ~\cite{yu2020searching} & 0.07 & 0.17 & 0.12 \\
      &DualStage ~\cite{wang2022disentangled} & 0.00 & 0.00 & \textbf{0.00} \\
      &NAS-FAS ~\cite{yu2020fas} & 0.07 & 0.17 & \underline{0.12} \\
      & \textbf{PatchNet (Ours)} & 0.00 & 0.00 & \textbf{0.00}\\
\hline
\multirow{5}{*}{2} 
      &Disentangle ~\cite{zhang2020face}& 0.08$\pm$0.17& 0.13$\pm$0.09& 0.10$\pm$0.04\\
      &SpoofTrace ~\cite{liu2020disentangling} & 0.00$\pm$0.00 &  0.00$\pm$0.00 & \textbf{0.00$\pm$0.00} \\
      &BCN ~\cite{yu2020face} & 0.08$\pm$0.17& 0.15$\pm$0.00 & 0.11$\pm$0.08 \\
      &CDCN ~\cite{yu2020searching}&0.00$\pm$0.00 &0.13$\pm$0.09  & 0.06$\pm$0.04 \\
      &DualStage ~\cite{wang2022disentangled} & 0.00$\pm$0.00 & 0.00$\pm$0.00 & \textbf{0.00$\pm$0.00}\\
      &NAS-FAS ~\cite{yu2020fas} &0.00$\pm$0.00 &0.09$\pm$0.10  & \underline{0.04$\pm$0.05} \\
      &\textbf{PatchNet (Ours)} & 0.00$\pm$0.00 & 0.00$\pm$0.00 & \textbf{0.00$\pm$0.00}\\
\hline
\multirow{5}{*}{3} 
      &Disentangle ~\cite{zhang2020face}& 9.35$\pm$6.14& 1.84$\pm$2.60& 5.59$\pm$4.37\\
      &SpoofTrace ~\cite{liu2020disentangling} & 8.3$\pm$3.3 & 7.5$\pm$3.3 & 7.9$\pm$3.3 \\
      &BCN ~\cite{yu2020face} & 2.55$\pm$0.89 & 2.34$\pm$0.47& 2.45$\pm$0.68\\
      &CDCN ~\cite{yu2020searching} & 1.67$\pm$0.11 & 1.76$\pm$0.12 & \underline{1.71$\pm$0.11} \\
      &DualStage ~\cite{wang2022disentangled} & 4.77$\pm$5.04 & 2.44$\pm$2.74 & 3.58$\pm$3.93 \\
      &NAS-FAS ~\cite{yu2020fas} & 1.58$\pm$0.23 & 1.46$\pm$0.08  & \textbf{1.52$\pm$0.13} \\
      &\textbf{PatchNet (Ours)} & 3.06$\pm$1.1 & 1.83$\pm$0.83 & 2.45$\pm$0.45\\
\bottomrule
\end{tabular}}
\caption{The results of testing on SiW protocols.}
\label{tab:SiW}
\vspace{-5.5mm}
\end{table}

\subsubsection{Results on SiW}
SiW~\cite{liu2018learning} is another commonly used high-quality dataset with more identities. The collection is captured by two different quality devices: Canon EOS T6 and Logitech C920. Compared to Oulu-NPU, it includes more environment variations and spoof mediums. The numbers of fine-grained patch type classes during training in protocol 1, 2, 3-1, and 3-2 are 14, 8, 6, and 10, respectively. As shown in Tab.~\ref{tab:SiW}, our method performs the best for the first two protocols and achieves competitive results in protocol 3.

\subsection{Ablation Study} \label{sec:ablation}
In this subsection, all ablation studies are conducted on Protocol 1 (different illumination conditions and location between the train and test sets) of OULU-NPU~\cite{boulkenafet2017oulu} to explore the details of our patch-based recognition framework. 

\begin{table}[t]
    \scriptsize
    \centering
    \resizebox{0.95\columnwidth}{!}{\begin{tabular}{cc|cc|cc|c}
        \toprule
        \multicolumn{2}{c|}{\textbf{Output Class}} & \multicolumn{2}{c|}{\textbf{Input Extraction}} & \multicolumn{2}{c|}{\textbf{Loss Functions}} & \multirow{2}{*}{\textbf{ACER(\%)}}\\
        \cline{1-6}
        Binary & Fine & Resize & PatchCrop & $\mathcal{L}_{Asym}$ & $\mathcal{L}_{Sim}$ & ~  \\
        \hline
        \hline
        \cmark &  & \cmark & & & & 6.25 \\
        \cmark &  &  &\cmark  & & & 3.54 \\
        &\cmark &  \cmark & & & & 5.63 \\
        &\cmark &   &\cmark  & & & 1.88 \\
        \hline
        &\cmark &   &\cmark  & & \cmark & 1.46 \\
        &\cmark &   &\cmark  & \cmark & & 0.63 \\
        &\cmark &   &\cmark  & \cmark & \cmark & 0.0 \\
        \bottomrule
    \end{tabular}}
    \caption{Ablation study of each component in PatchNet on OULU-NPU protocol 1.}
    \label{tab:ablation}
    \vspace{-4.5mm}
\end{table}

\begin{figure}
\begin{minipage}[b]{.48\linewidth}
    \includegraphics[width=1.0\linewidth]{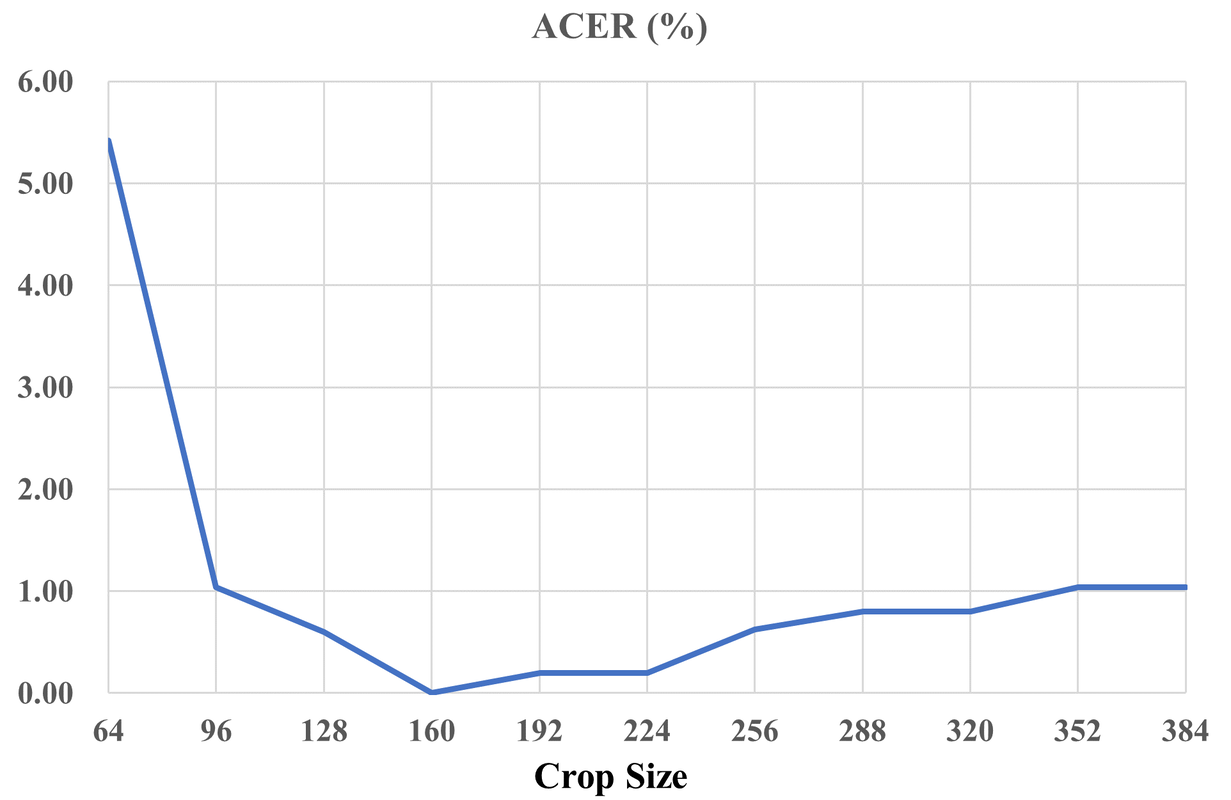}
    \caption{Comparison between choices of patch crop size.}
    \label{fig:crop-size}
\end{minipage}
\hspace{1mm}
\begin{minipage}[b]{.48\linewidth}
    \centering
    \scriptsize
    \begin{tabular}{c|c|c}
    \toprule
    $m_l$ & $m_s$ & ACER(\%) \\
    \hline
    \hline
    0.0 & 0.0 & 1.46 \\
    0.2 & 0.2 & 0.83 \\
    0.2 & 0.1 & 0.41 \\
    0.4 & 0.4 & 0.63 \\
    0.4 & 0.3 & 0.2 \\
    0.4 & 0.2 & 0.2 \\
    \textbf{0.4} & \textbf{0.1} & \textbf{0.0} \\
    0.4 & 0.0 & 0.41 \\
    \bottomrule
    \end{tabular}
    \captionof{table}{Ablation study of margin choices in $L_{Asym}$.}
    \label{tab:compare1}
\end{minipage}
\vspace{-3.5mm}
\end{figure}

\textbf{Impact of Each Component.} Tab.~\ref{tab:ablation} shows the ablation study of each component in our proposed framework. The first row is the naive baseline (ACER: 6.25\%) which formulates the face anti-spoofing as a binary classification problem (trained with the standard Cross Entropy loss) with the resized 256x256 face input. Surprisingly, by only adopting the fine-grained classes and raw frame cropping strategies, we can improve the performance significantly to 1.88\% ACER. It shows that the naive baseline model could overfit to the high-level biases in the public FAS dataset, which only contains limited background and identities. Moreover, the fine details in cropped patches from raw frames are very critical to discriminate between different patch types in high-quality datasets like OULU-NPU. From the lower part of the table, we can observe the advantage of the proposed margin-based classification loss and the self-supervised similarity loss. It is clear that both regularization techniques can facilitate the encoder to learn more intrinsic features related to the capture device's characteristics and presenting materials. 

\textbf{Impact of Patch Crop Size.} Fig.~\ref{fig:crop-size} demonstrates the ACER(\%) on OULU-NPU protocol 1 between different crop sizes. We can observe that larger patch sizes during training might be prone to overfit to biases from the face capture. With the regularization of patch recognition loss and patch-based augmentation to increase training data variance, the overall performance does not differ much when enlarging the patch size. However, when the patch size is too small (e.g., 64), the performance degrades significantly as the capture characteristics can not be learned with very limited information.

\textbf{Asymmetric Margin Choices.} We conduct ablation experiments to verify the effectiveness of our asymmetric margin design in the angular margin softmax loss. From Tab.~\ref{tab:compare1}, we can observe that adding angular margins can significantly improve the generalization capability and outperform the model without any margin. Furthermore, adding a very large margin to spoof patch types which have more diverse appearances would hurt the discrimination power of learned features. We find that PatchNet works well on all testing protocols with the margin choice $m_l = 0.4$, $m_s = 0.1$.

\subsection{Cross-Dataset Testing} \label{sec:inter-results}
\subsubsection{Experiments between C and I} 
First, following the related works, CASIA-MFSD (C)~\cite{zhang2012casia} and ReplayAttack (I)~\cite{chingovska2012replayattack} are used for cross-dataset experiments, and the results are measured in HTER. During training, the numbers of fine-grained patch-type classes are 9 and 4 in $C \rightarrow I$ and $I \rightarrow C$ protocols, respectively. The results are shown in Tab.~\ref{tab:cross-testing}. Given the limited number of clips and low-quality videos from the ReplayAttack dataset, it is hard to learn generalizable features which can perform very well on other datasets, so the error rate in protocol $I \rightarrow C$ is still high compared with $C \rightarrow I$. Our proposed framework can achieve competitive performance compared to previous works in both protocols.      

\newcommand{\tabincell}[2]{\begin{tabular}{@{}#1@{}}#2\end{tabular}}
\begin{table}
\centering
\resizebox{0.95\columnwidth}{!}{\begin{tabular}{c|c|c|c|c}
    \toprule
    \multirow{2}{*}{Method} &Train &Test &Train &Test\\
    \cline{2-3} \cline{4-5} &\tabincell{c}{CASIA-\\MFSD(C)} &\tabincell{c}{Replay-\\Attack(I)} &\tabincell{c}{Replay-\\Attack(I)} &\tabincell{c}{CASIA-\\MFSD(C)}\\
    \hline
    \hline
    STASN ~\cite{yang2019face}
    &\multicolumn{2}{c|}{31.5} &\multicolumn{2}{c}{30.9} \\
    Disentangle ~\cite{zhang2020face}
    &\multicolumn{2}{c|}{22.4} &\multicolumn{2}{c}{30.3} \\
    BCN ~\cite{yu2020face}
    &\multicolumn{2}{c|}{16.6} &\multicolumn{2}{c}{36.4} \\
    CDCN ~\cite{yu2020searching}
    &\multicolumn{2}{c|}{15.5} &\multicolumn{2}{c}{32.6} \\
    DC-CDN ~\cite{yu2021dual}
    &\multicolumn{2}{c|}{\textbf{6.0}} &\multicolumn{2}{c}{\underline{30.1}} \\
    \textbf{PatchNet (Ours)}
    &\multicolumn{2}{c|}{\underline{9.9}} &\multicolumn{2}{c}{\textbf{26.2}} \\
    \bottomrule
\end{tabular}}
\caption{The results of cross-dataset testing between CASIA-MFSD and Replay-Attack. The evaluation metric is HTER(\%).}
\label{tab:cross-testing}
\vspace{-3.5mm}
\end{table}

\begin{table*}[]
    \centering
    \resizebox{0.8\textwidth}{!}{\begin{tabular}{c|c|c|c|c|c|c|c|c}
        \toprule
        \multirow{2}*{\textbf{Method}} & \multicolumn{2}{c|}{\textbf{O\&C\&I to M}} & \multicolumn{2}{c|}{\textbf{O\&M\&I to C}} & \multicolumn{2}{c|}{\textbf{O\&C\&M to I}} & \multicolumn{2}{c}{\textbf{I\&C\&M to O}} \\
        \cline{2-9}
        ~ & HTER(\%) & AUC(\%) & HTER(\%) & AUC(\%) & HTER(\%) & AUC(\%) & HTER(\%) & AUC(\%)  \\
        \hline
        \hline
        PatchNet w/ coarse cls~\cite{jia2020single} & 10.24 & 96.45 & 15.67 & 92.47 & 21.65 & 91.08 & 16.26 & 91.33 \\
        \hline
        PatchNet w/o margin & 10.0 & 96.61 & 18.0 & 91.57 & 17.25 & 90.47 & 15.04 & 92.42 \\
        PatchNet w/o $\mathcal{L}_{Sim}$ & 8.9 & 97.42 & 13.44 & 93.99 & 15.1 & 92.10 & 14.24 & 92.93 \\
        \hline
        \textbf{PatchNet (Ours)} & \textbf{7.10} & \textbf{98.46} & \textbf{11.33} & \textbf{94.58} & \textbf{14.6} & \textbf{92.51} & \textbf{11.82} & \textbf{95.07} \\
        \bottomrule
    \end{tabular}}
    \caption{Evaluations of different components of the proposed method on four cross-dataset protocols.}
    \label{tab:dg_ablation}
    \vspace{-3.5mm}
\end{table*}

\begin{table*}[]
    \centering
    \resizebox{0.8\textwidth}{!}{\begin{tabular}{c|c|c|c|c|c|c|c|c}
        \toprule
        \multirow{2}*{\textbf{Method}} & \multicolumn{2}{c|}{\textbf{O\&C\&I to M}} & \multicolumn{2}{c|}{\textbf{O\&M\&I to C}} & \multicolumn{2}{c|}{\textbf{O\&C\&M to I}} & \multicolumn{2}{c}{\textbf{I\&C\&M to O}} \\
        \cline{2-9}
        ~ & HTER(\%) & AUC(\%) & HTER(\%) & AUC(\%) & HTER(\%) & AUC(\%) & HTER(\%) & AUC(\%)  \\
        \hline
        \hline
        Auxiliary \cite{liu2018learning} & 22.72 & 85.88 & 33.52 & 73.15 & 29.14 & 71.69 & 30.17 & 77.61 \\
        MADDG \cite{shao2019multi}  & 17.69 & 88.06 & 24.50 & 84.51 & 22.19 & 84.99 & 27.89 & 80.02 \\
        PAD-GAN \cite{wang2020cross} & 17.02 & 90.10 & 19.68 & 87.43 & 20.87 & 86.72 & 25.02 & 81.47 \\
        RFM \cite{shao2020regularized} & 13.89 & 93.98 & 20.27 & 88.16 & 17.30 & 90.48 & 16.45 & 91.16 \\
        NAS-FAS \cite{yu2020fas} & 16.85 & 90.42 & 15.21 & 92.64 & \textbf{11.63} & \textbf{96.98} & 13.16 & 94.18 \\
        SSDG-R \cite{jia2020single} &7.38& 97.17& 10.44& 95.94& 11.71 &96.59& 15.61& 91.54\\
        ANRL ~\cite{liu2021adaptive} &16.03& 91.04& \textbf{10.83}& \textbf{96.75}& 17.85& 89.26& 15.67& 91.90 \\
        DRDG ~\cite{liu2021dual} &15.56& 91.79& 12.43& 95.81& 19.05& 88.79& 15.63& 91.75 \\ 
        \hline
        \textbf{PatchNet (Ours)} & \textbf{7.10} & \textbf{98.46} & 11.33 & 94.58 & 13.4 & 95.67 & \textbf{11.82} & \textbf{95.07} \\
        \bottomrule
    \end{tabular}}
    \caption{Comparison results between the proposed PatchNet and state-of-the-art methods on four domain generalization protocols.}
    \label{tab:dg_sota}
    \vspace{-3.5mm}
\end{table*}
\vspace{-1.5mm}
\subsubsection{Domain Generalization Experiments}
Some recent FAS works~\cite{wang2020cross, jia2020single, shao2020regularized} consider each dataset as one domain and promote the domain generalization benchmark in FAS, which utilizes three datasets for training, and the remaining one as testing. As we aim to distinguish the patch type in the fine-grained manner, our proposed framework can be directly used to evaluate such benchmark without employing further generalization techniques (e.g., adversarial training or meta-learning). With access to more different patch types with more diverse capture devices, our framework is capable of learning more discriminative features through the patch recognition proxy task. There are four protocols in this benchmark: O\&C\&I to M, M\&I\&O to C, M\&C\&O to I, and M\&C\&I to O. During training, we directly combine the fine-grained patch classes from the three training datasets, which results in 18, 17, 22, and 21 classes, respectively.   

The testing results are shown in Tab.~\ref{tab:dg_sota}. The proposed PatchNet achieves competitive results on all protocols. Due to the high variance of capture types in C dataset, it is hard to learn robust local features to address both high and low resolution scenarios. We also conduct an ablation study in this benchmark to explore the influence of each component in our framework and show the results in Tab.~\ref{tab:dg_ablation}. In the first ablation experiment, we split the patch classes using the strategy proposed by SSDG~\cite{jia2020single}: It aggregates the live samples as one class and treats spoof samples from each other dataset as one class, which results in 4 classes. The results verify that the proposed fine-grained class split, $L_{Asym}$, and $L_{Sim}$ are all important to regularize the network to generalize better in challenging FAS tasks.      

\subsection{Visualizations} \label{sec:visualization}
\begin{figure}[t]
    \centering
    \includegraphics[width=.95\linewidth]{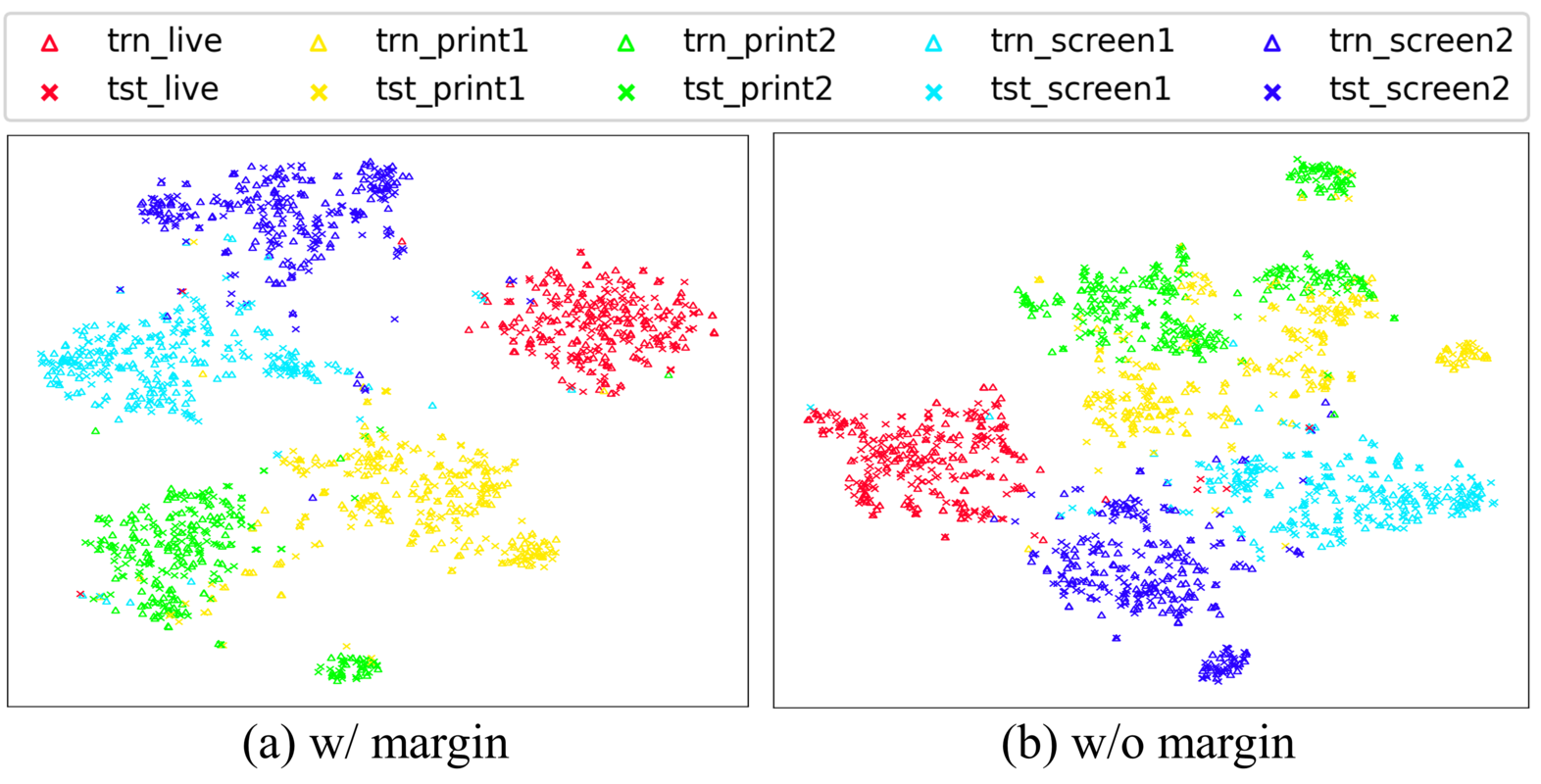}
    \caption{The t-SNE features visualizations on Oulu-NPU Protocol 1. (a) training with asymmetric margin (b) training without margin.}
    \label{fig:tsne}
    \vspace{-3.5mm}
\end{figure}

\textbf{Patch Feature Distribution.} In Fig.~\ref{fig:tsne}, we visualize the patch features by t-SNE~\cite{van2008visualizing} from OULU-NPU protocol 1, which both the training and testing sets consist of 5 patch types (live, print1, print2, screen1, screen2). We can observe in (b) that the model trained without margin cannot distinguish print1 and print2 types very well. The distribution for the live samples is more compact, and clusters are separated better for the model trained with the margin. The training and testing feature sets are aligned well in the feature space.  

\begin{figure}[t]
    \centering
    \includegraphics[width=.95\linewidth]{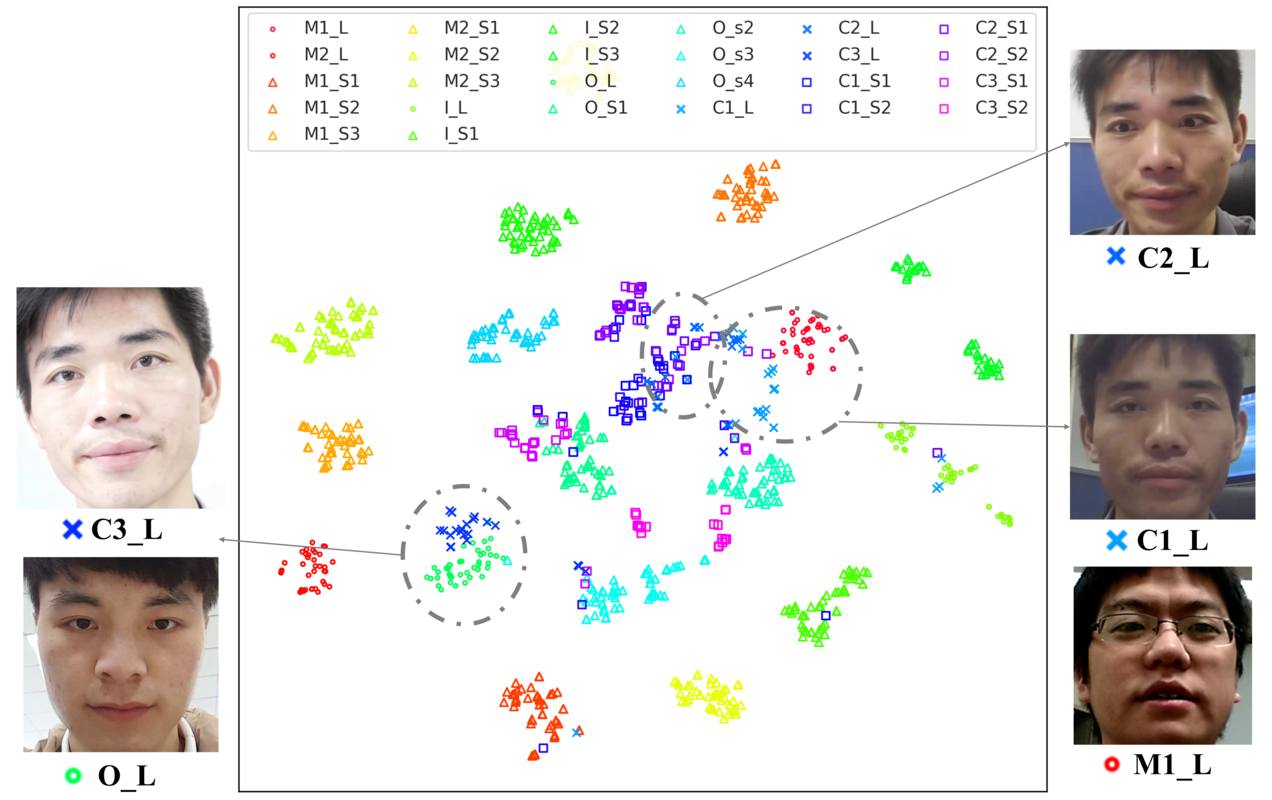}
    \caption{The t-SNE visualizations of normalized patch features in cross-dataset protocol M\&I\&O to C. The fine-grained patch type class is denoted by (Dataset)(SensorID)\_(Liveness)(MediumID).}
    \label{fig:tsne2}
    \vspace{-3.5mm}
\end{figure}

\begin{figure}[t]
    \centering
    \includegraphics[width=.95\linewidth]{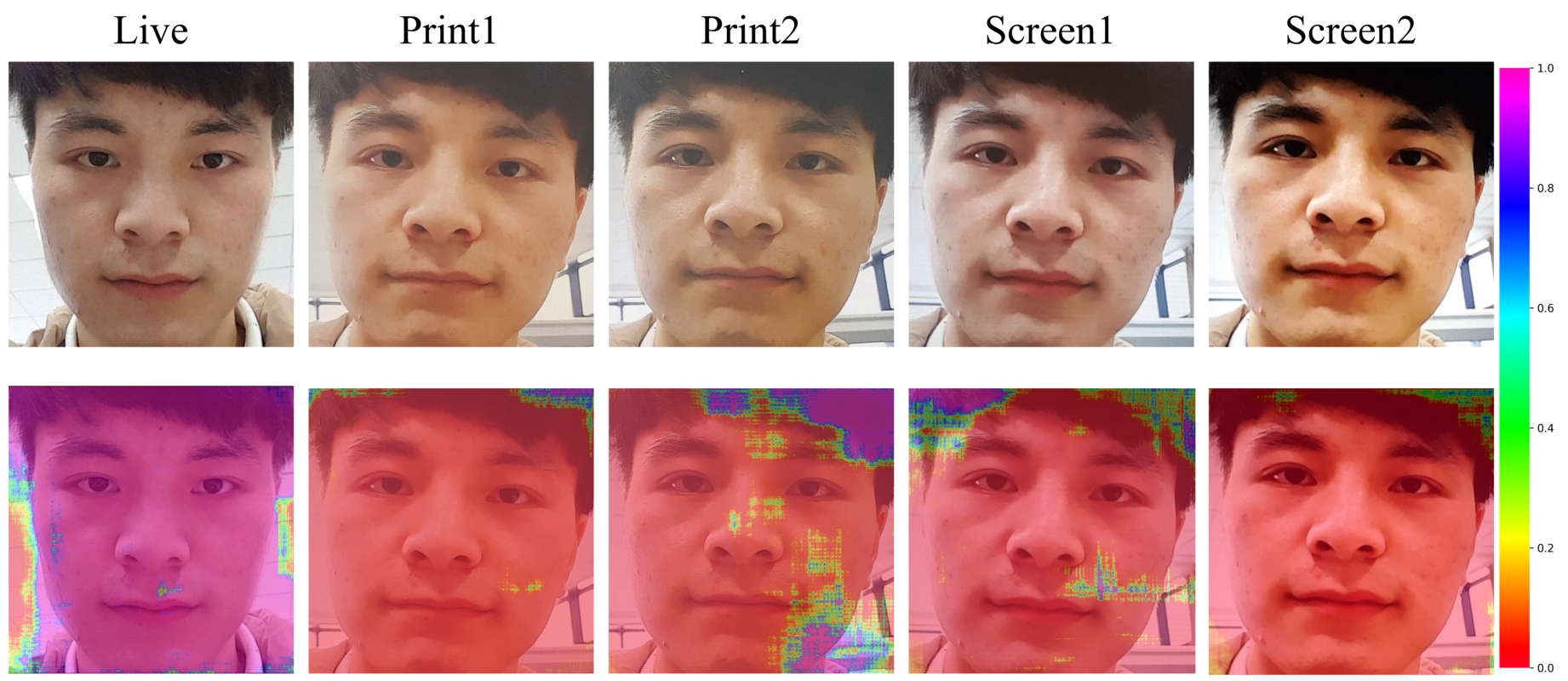}
    \caption{Patch score map of 5 different types in Oulu-NPU Protocol 1. From left to right: Live, Print1, Print2, Screen1, and Screen2. The number represents the live probability.}
    \label{fig:patch_score}
    \vspace{-3.5mm}
\end{figure}

In Fig.~\ref{fig:tsne2}, we visualize the feature distribution in the protocol M\&I\&O to C. We can observe that the features from the training set are separated well in the fine-grained manner. As dataset C contains three different quality capture devices, the corresponding three live classes are located separately in the embedding space. Aligning with the device characteristics, we observe that features from C3\_L are close to O\_L, which both types are captured from high-quality capture devices. Moreover, low-quality devices like C1\_L exhibit similar appearance with that from similar quality device M1\_L. It implies that the patch embedding space encodes the capture characteristic well.  

\textbf{Patch Score Map.} In Fig.~\ref{fig:patch_score}, we compute the scores from patches across the whole image and visualize the live probability in the overlapped heat map. The samples are from the testing set of OULU-NPU protocol 1. The patch scores across the whole face input image are primarily consistent, except for some background and boundary parts, which is expected as the spoof cues should be learned from the face region but not the background's biases.   

\subsection{Applications} \label{sec:few-shot}
\textbf{Testing on each Capture Device.} Following the observation in Sec.~\ref{sec:visualization}, we argue that instead of testing on the whole dataset, it is more feasible to test the anti-spoofing performance on each device separately (e.g., in the CASIA-FASD dataset). We re-organize the testing dataset in O\&C\&I to M and O\&M\&I to C protocols, which is splitting the testing data by their capture device ID into 2 (M1, M2) and 3 (C1, C2, C3) groups, respectively. The cross-dataset anti-spoofing performance on the new split dataset of these two protocols is shown in Tab.~\ref{tab:split-test}. We observe that anti-spoofing performance on both devices in the M dataset are better than the average performance using the whole dataset, which is reasonable as the testing sets are smaller. However, only one device (the high-quality one C3) in the C dataset is better than the average performance, which aligns with the t-SNE visualization in Fig.~\ref{fig:tsne2}. The capture images of C2 have many noises and image compression effects, which can lead to significant degradation of the discriminative power of features. With the detailed performance report on each device, we can further improve the anti-spoofing system or improve the quality of the problematic capture devices as well.  

\textbf{Few-Shot Reference Anti-Spoofing.} With the learned patch embedding space which encodes intrinsic patch features to discriminate between patch types, the distance between features in the embedding space can be used to measure the similarity between patch types. As the features are already normalized in the space, we can compute the cosine distance between features to enable a new application: few-shot reference anti-spoofing. While there is a new capture device, it is easier to acquire some \textbf{live} face image samples, and those sample features can be used as the reference in the embedding space. We compute the distances between other testing features and live reference to obtain the similarity scores. Higher similarity scores mean the samples are more likely to be live. The 5-shot and 10-shot live reference anti-spoofing performance are reported in Tab.~\ref{tab:split-test}. We can conclude that with the live reference features and the learned patch embedding space, the performance can be boosted a lot in practical scenarios.       

\begin{table}[t]
    \centering
    \resizebox{0.95\columnwidth}{!}{\begin{tabular}{c|c|c|c|c|c}
        \toprule
        \multirow{3}*{\textbf{Method}} & \multicolumn{2}{c|}{\textbf{O\&C\&I to M}} & \multicolumn{3}{c}{\textbf{O\&M\&I to C}} \\
        & \multicolumn{2}{c|}{(AUC: 98.46\%)} & \multicolumn{3}{c}{(AUC: 94.58\%)} \\
        \cline{2-6}
        & M1 & M2 & C1 & C2 & C3\\
        \hline
        \hline
        PatchNet & 99.54 & 98.63 & 94.29 & 88.49 & 98.13 \\
        \hline
        PatchNet w/ 5-shot & 99.7 & 99.6 & 94.3 & 89.8 & 98.6 \\
        PatchNet w/ 10-shot & 99.8 & 99.6 & 95.3 & 90.7 & 99.2 \\
        \bottomrule
    \end{tabular}}
    \caption{Split testing and few-shot live reference testing on each capture device. The few-shot testing score is averaged by 10 experiment runs. AUC(\%) score is reported.}
    \label{tab:split-test}
    \vspace{-3.5mm}
\end{table}
\begin{figure}[t]
    \centering
    \includegraphics[width=.98\linewidth]{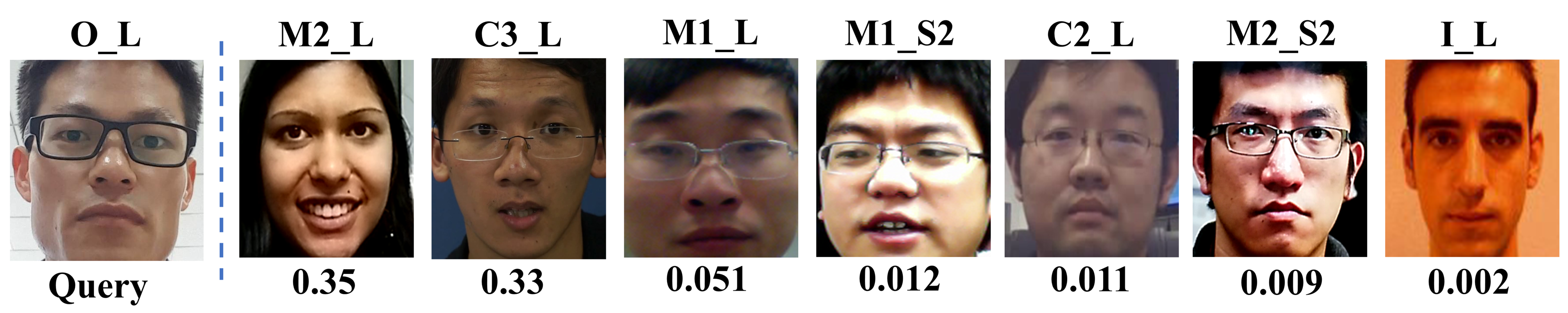}
    \caption{Patch type retrieval using a live sample from O dataset as the query. Top retrieval results are shown, and the numbers below are the cosine similarity.}
    \label{fig:retrieval}
    \vspace{-4.5mm}
\end{figure}
\textbf{Patch Type Retrieval.} \label{sec:retrieval}
The normalized patch embedding space can be used for the patch type retrieval application, and can boost the FAS performance on certain capture devices. For example, after training the feature space with M\&C\&I datasets to recognize 21 patch types, we can measure the similarity between testing patch types and training patch types. As shown in Fig.~\ref{fig:retrieval}, we take a live sample from O dataset as the query, and retrieve the training patch types by computing the cosine distance with each type weight vector. The top-7 retrieval patch types are (M2\_L, C3\_L, M1\_L, M1\_S2, C2\_L, M2\_S2, I\_L). Some spoof types from M dataset have higher ranking than live types from C2 and I datasets, which have low quality captures. To optimize the performance while testing on O dataset, we can 1) remove ambiguous live types (C2\_L, C1\_L, I\_L) during testing, or 2) re-define the class for (C2\_L, C1\_L, I\_L) as ``spoof" for training. Both strategies can boost the performance to 95.27\% and 95.87\% AUC, respectively.

%% file: 5_conclusions.tex
\section{Conclusions and Future Work}
In this paper, we reformulate face anti-spoofing as a fine-grained patch type recognition task and present a simple training framework called PatchNet to efficiently learn the patch embedding space which encodes the spoof-related capture characteristics. The novel loss functions are designed to enhance the feature discrimination power. Extensive experiments on challenging FAS protocols verify the effectiveness of the proposed method. We note that exploration of a generic embedding space to discriminate different captures is still at an early stage. Future directions include: 1) Learning a more generalized embedding space by datasets with more variations or transferred from the material perception task, and 2) Investigation into the Few-Shot FAS protocols which have more practical values.